\documentclass{ecai}

\usepackage[T1]{fontenc} 
\usepackage[utf8]{inputenc}

\usepackage{times}
\usepackage{graphicx}
\usepackage{latexsym}
\usepackage{booktabs}
\usepackage{amsmath}
\usepackage{amsfonts}

\usepackage[firstpageonly,angle=0,fontsize=1em,vpos=2em,color={[gray]{0.5}},text={ECAI 2020. Author version. Final manuscript is published online by IOS Press: {doi.org/10.3233/FAIA200075}}]{draftwatermark}

\begin{document}

\title{On-Time Delivery in Crowdshipping Systems: An Agent-Based Approach Using Streaming Data}

\author{Jeremias Dötterl\institute{Hannover University of Applied Sciences and Arts, Germany\newline jeremias.doetterl@hs-hannover.de}
    \and Ralf Bruns\institute{Hannover University of Applied Sciences and Arts, Germany\newline ralf.bruns@hs-hannover.de}
    \and Jürgen Dunkel\institute{Hannover University of Applied Sciences and Arts, Germany\newline juergen.dunkel@hs-hannover.de}
    \and Sascha Ossowski\institute{CETINIA, Universidad Rey Juan Carlos, Madrid, Spain\newline sascha.ossowski@urjc.es}
}

\maketitle

\begin{abstract}
In parcel delivery, the ``last mile'' from the parcel hub to the customer is costly, especially for time-sensitive delivery tasks that have to be completed within hours after arrival.
Recently, crowdshipping has attracted increased attention as a new alternative to traditional delivery modes.
In crowdshipping, private citizens (``the crowd'') perform short detours in their daily lives to contribute to parcel delivery in exchange for small incentives.
However, achieving desirable crowd behavior is challenging as the crowd is highly dynamic and consists of autonomous, self-interested individuals.
Leveraging crowdshipping for time-sensitive deliveries remains an open challenge.
In this paper, we present an agent-based approach to on-time parcel delivery with crowds.
Our system performs data stream processing on the couriers' smartphone sensor data to predict delivery delays.
Whenever a delay is predicted, the system attempts to forge an agreement for transferring the parcel from the current deliverer to a more promising courier nearby.
Our experiments show that through accurate delay predictions and purposeful task transfers many delays can be prevented that would occur without our approach.
\end{abstract}

\section{Introduction}
The ``last mile''~\cite{lee2001} is known to be one of the most costly and challenging elements of parcel delivery~\cite{gevaers2014,savelsbergh2016}.
Deliveries from the parcel delivery company's hub to the customers' private homes traditionally require a fleet of professional drivers who perform dedicated trips.
As each driver is responsible for many parcels, they lack the flexibility to satisfy time-sensitive delivery requests that arrive dynamically over the course of the day.
With ongoing e-commerce growth and an elevated desire for high delivery speeds, the challenges of the last mile are expected to increase even more in the future~\cite{savelsbergh2016}.

Crowdshipping~\cite{sadilek2013} is a recent trend of urban logistics that leverages ``the crowd'' for on-demand parcel delivery.
Private citizens can register at the service and gain monetary rewards for delivering parcels.
The core idea is that participants make short detours in their daily commutes, e.g., on their way to the supermarket.
As the crowd consists of many individuals who can respond to delivery requests, dynamically arriving delivery tasks can be performed in parallel whereas professional drivers would perform them sequentially~\cite{wang2016}.
As short detours require little monetary compensation and can be completed rapidly, crowdshipping holds the potential to become a low-priced option for time-sensitive delivery in urban areas.

To fulfill this potential and to gain consumer acceptance, crowdshipping has to be reliable.
Consumers are dissatisfied if their deliveries do not arrive on time~\cite{visser2014}.
Therefore, it is essential that the crowd is able to provide on-time delivery.
Current research on crowdshipping focuses mostly on the initial assignment of couriers to delivery tasks and on upfront route planning \cite{archetti2016,arslan2019,chen2018,dayarian2017,gdowska2018,wang2016}, but tends to neglect delivery execution:
We argue that, to achieve on-time delivery in crowdshipping systems, delivery execution has to be monitored and if problems arise, corrective measures have to be taken.

In this paper, we present an agent-based approach~\cite{wooldridge1995} to reduce delivery delays in crowdshipping systems.
The agent-based system performs data stream mining \cite{gama2010} on the couriers' smartphone sensor data to predict delivery delays.
Whenever a delay is predicted, the system tries to forge an agreement for transferring the delivery task from the current deliverer to another courier who is willing and capable of completing the delivery in time.
Transfers allow tasks to ``move within the crowd'' from slow couriers to faster couriers who, e.g., use another transportation mode.
Our approach treats the couriers as autonomous agents who only agree to perform delivery tasks when they are sufficiently compensated.

Our paper makes the following contributions:
\begin{itemize}
    \item We present a crowdshipping monitoring approach that learns to predict delivery delays from the crowd's smartphone sensor data.
    \item We present a transfer approach that forges transfer agreements between couriers based on the delay predictions.
        The transfer approach respects the autonomy of the couriers.
\end{itemize}

The rest of this paper is structured as follows.
First, we discuss the most relevant related work on crowdshipping (Section~\ref{sec:related-work}).
Then, we introduce the problem of parcel delivery with autonomous couriers (Section~\ref{sec:problem}).
In Section~\ref{sec:approach}, we propose an agent-based crowdshipping architecture for on-time parcel delivery.
In Section~\ref{sec:datastreams}, we explain our use of data stream processing to derive delay predictions from smartphone sensor data.
How the system forges transfer agreements between the couriers based on the delay predictions is explained in Section~\ref{sec:agreements}.
In Section~\ref{sec:evaluation}, we evaluate our approach with experiments using real GPS data gathered from a real crowd (the bike sharing users of Madrid) and with a crowdshipping simulation.
Finally, we close with concluding remarks (Section~\ref{sec:conclusion}).

\section{Related Work}
\label{sec:related-work}

Recently, there has been increased interest in crowdsourced delivery.
Besides research on the users' acceptance to use or participate in a crowdshipping system \cite{marcucci2017,punel2018,punel2017}, many works focus on the (initial) task-to-courier assignment and/or route planning.

Different assumptions can be made about a courier's willingness to accept a delivery task.
Sometimes a threshold for the detour is used and the couriers are willing to accept a delivery task if the required detour does not exceed a certain fraction of the length of their original route \cite{archetti2016,chen2018}.
Similarly, some authors assume that each courier has a coverage area (e.g., within a certain radius of the courier's destination) and accepts any task within this area \cite{dayarian2017}.
Besides spatial considerations, temporal aspects can be included: In \cite{arslan2019} the couriers accept any task that satisfies temporal constraints such as the latest arrival time, maximum travel time, and departure time flexibility.
In \cite{chen2017} a detour time threshold is used.
Other authors do not model the couriers' decision making explicitly and the couriers accept and reject tasks stochastically with a certain probability \cite{gdowska2018}.
Couriers can also be modeled as self-interested, autonomous agents that accept delivery tasks if they provide sufficient utility \cite{angel2018}, where the utility is influenced by the delivery reward and costs (fuel, vehicle maintenance, delivery distance, etc.).
In this paper, we adopt such a utility-based perspective and model couriers as autonomous agents that consider delivery costs, rewards, deadlines, and penalties for deadline violations.

In the literature, different assumptions are made about the knowledge that is available about the individual couriers.
Some methods require knowledge about a courier's usual travel pattern to compute appropriate assignments \cite{wang2016}.
Other methods require that couriers report their temporal availability, to be able to compute assignments and routes that satisfy these temporal constraints \cite{arslan2019}.
In static settings, it is assumed that all delivery tasks and couriers are known upfront \cite{chen2018}.
In this paper, we use smartphone sensor data to monitor delivery execution after the assignment to prevent delivery delays.

Few approaches actively influence the delivery outcome by considering task transfers.
In \cite{angel2018}, transfers are performed when two drivers benefit financially and redundant trips can be avoided.
In \cite{chen2018}, parcels are exchanged between couriers, but according to a plan that is computed offline.
In \cite{giret2018b,rebollo2018} parcels are exchanged between couriers based on live information about the city's transportation network and the couriers' GPS locations.

Some related work considers temporal aspects in crowdshipping, e.g., the couriers' temporal availability~\cite{chen2017crowddeliver,chen2018}.
Until today, there is little research on crowdshipping systems that actively monitor the delivery to prevent delays.
The work with the most similar aim that we have found is by Habault et al.~\cite{habault2018}, who present a delivery management system for last-mile (food) delivery.
The system monitors the delivery vehicles to provide route planning.
The authors propose to use Machine Learning to predict traffic to recommend faster routes and thus to reduce delivery delays.
While this work pursues a similar goal as ours, there are great differences in our approach:
We are not aware of any related work that combines agent-based systems with data stream mining and that actively prevents delivery delays through task transfers.

\section{Parcel Delivery with Autonomous Couriers}
\label{sec:problem}

In this section, we introduce the crowdshipping setting to which our solution approach (Sections~\ref{sec:approach} to \ref{sec:agreements}) can be applied.
We pay special attention to the decision making of the autonomous couriers, which motivates the need for transfer \emph{agreements}.
This decision making model is later used in our agent-based simulation to evaluate the effectiveness of our proposal (Section~\ref{sec:evaluation-delay-prevention}).

The crowdshipping system operates in a restricted geographical area where point-to-point delivery tasks occur, e.g., in the city of Madrid.
The system bridges the last mile: parcels are shipped between packing stations and the customers' home addresses (in both directions).
A parcel delivery task $p = (l_p, d_p, \tau_p, r_p, s_p)$ is characterized by its start location $l_p$ and its destination $d_p$.
If the parcel arrives at its destination before the deadline $\tau_p$, the deliverer receives a monetary reward $r_p$.
If the parcel arrives after the deadline, the deliverer is sanctioned and has to pay a penalty $s_p$.

The delivery tasks are initially assigned to the couriers such that all couriers accept their assignment, e.g., via auction-based assignment.
A courier $i \in \mathcal{A}$ accepts a task if the courier believes that they can complete the task before the deadline and if the required detour from the courier's initially intended route is sufficiently compensated.
A courier $i$ is characterized by a tuple $(l_i, d_i, \mathit{U}_i)$ where
$l_i$ is the courier's location,
$d_i$ the courier's destination,
and $\mathit{U}_i$ the courier's utility function.
The function $\mathit{U}_i$ specifies the utility that courier $i$ derives from the assignment $(i, p)$:
$$\mathit{U}_i(i, p) =
\begin{cases}
r_p - \mathit{C}_i(p) & \text{if } t_{i,p} < \tau_p \\
-s_p - \mathit{C}_i(p) & \text{otherwise. }
\end{cases}$$
where $t_{i,p}$ denotes the time at which $i$ finishes $p$ and
$\mathit{C}_i(p)$ denotes the courier's cost (fuel, vehicle maintenance, effort, etc.) for $p$.
The value of $\mathit{C}_i(p)$ depends on the detour that the courier has to make from their initially intended route from $l_i$ to $d_i$.
The detour is the length of the path $l_i \rightarrow l_p \rightarrow d_p \rightarrow d_i$ minus the length of the courier's original path $l_i \rightarrow d_i$.
For simplicity, we assume that each courier accepts at most a single task at a time.
We assume that the couriers can perform a rough estimation of arrival time $t_{i,p}$.
Their estimated arrival time $\hat{t}_{i,p} = t_{i,p} + e$ where $e$ is a random error value from an interval $[-E, +E]$.
Hence, couriers can roughly estimate their utility but not precisely due to the noise $e$ in their arrival predictions.
A courier $i$ accepts a task $p$ if their estimated utility is positive.
We call a courier who has currently a task assigned \emph{deliverer}.

Ideally, the deliverer finishes the delivery on time and collects the promised reward.
However, parcel delivery takes place in an urban environment with many sources of disruption that can prevent the timely delivery.
Technical difficulties like car breakdowns or private and work-related interferences can make the deliverer unable or unwilling to complete the task before the specified deadline.
It is in the interest of both the logistics provider and the deliverer that delays are prevented.

\section{Delay Prevention through Proactive Task Transfers}
\label{sec:approach}

\subsection{General Idea}

The general idea of our approach is to anticipate delivery delays and to prevent their occurrence by transferring the delivery task to another, more promising courier, who we call the \emph{substitute}.
Through the transfer, the substitute assumes the responsibility of the delivery task:
if the parcel arrives on time, the substitute receives the full parcel reward $r_p$; if the parcel arrives with delay, the substitute has to pay penalty $s_p$.
The substitute would accept the task if they believe that they can complete the delivery in time and if the task provides a sufficient utility.
The deliverer would give away the task to avoid an anticipated penalty or if the substitute is willing to purchase the task from the deliverer for a sufficiently high price.
We discuss the details of this transaction in Section~\ref{sec:agreements}.

\subsection{Agent-Based Crowdshipping Architecture}
In this section, we present the high-level architecture of our approach, which gives an overview of the main actors and components and shows how they interact.
\begin{figure*}
    \centering
    \includegraphics[width=1.0\linewidth]{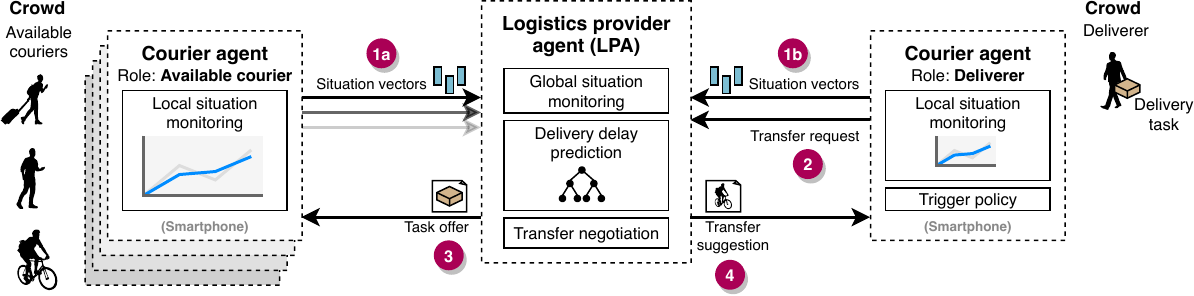}
    \caption{Agent-based crowdshipping architecture}
    \label{fig:architecture}
\end{figure*}
The architecture is shown in Fig.~\ref{fig:architecture}, which we keep referring to throughout this section.
As shown in the figure, the crowd consists of couriers who are represented in our system by \emph{courier agents}, which run on the couriers' smartphones.
The couriers can adopt two different roles:
\begin{itemize}
    \item \emph{Available couriers} are participants of the crowdshipping system who are available to perform a delivery task, but have no task assigned.
    \item \emph{Deliverers} are currently performing a delivery task that has been assigned to them.
\end{itemize}
The interests of the logistics provider are represented by the \emph{logistics provider agent} (LPA).
The architecture consists of the following components:
\begin{itemize}
\item \emph{Local situation monitoring}:
Independent of the courier's role, the courier agents monitor the couriers' local situations.
The courier agents perform data stream processing on the smartphone sensor data to derive situation vectors.
Situation vectors characterize a courier's live situation with information such as the courier's location, speed, traveling direction, estimated transportation mode, etc.
These situation vectors are periodically transmitted to the LPA, as indicated by the arrows 1a and 1b in the figure.

\item \emph{Global situation monitoring}:
The LPA gathers the situation vectors to monitor the global situation to maintain an up-to-date view onto the crowdshipping system.

\item \emph{Trigger policy}:
For couriers who play the deliverer role, the courier agent uses the situation vectors to detect delivery incidents.
The courier agent has a trigger policy which decides, based on the latest situation vector, whether the courier's delivery task should be transferred to another courier.
If the trigger policy fires, a transfer request is sent to the LPA (arrow 2).

\item \emph{Delivery delay prediction}:
The LPA uses situation vectors for delivery delay prediction.
Given a situation vector and a delivery task, the LPA estimates the likelihood that the courier who is in that situation can/could terminate the delivery task on time.

\item \emph{Transfer negotiation}:
The delay predictions are used to forge transfer agreements.
When the LPA receives a transfer request from a courier agent, it uses the delay predictions to find a capable substitute who could successfully continue the task.
Once a promising candidate is found, the LPA starts the transfer negotiation by offering the task to the candidate (arrow 3).
If the candidate declares interest in the task, the LPA presents the candidate to the deliverer (arrow 4), who can decide whether they want to transfer their task to this candidate.
If an agreement can be reached, the candidate becomes the substitute.
Otherwise, a transfer has to be attempted with another candidate (step 3 and 4 have to be repeated) or at a later point of time.
\end{itemize}

The next two sections provide more detailed descriptions of the six components.
\begin{itemize}
    \item Section~\ref{sec:datastreams} covers the five components related to data stream processing on smartphone sensor data: local situation monitoring, global situation monitoring, the trigger policy, and delivery delay prediction.
    \item Section~\ref{sec:agreements} covers the transfer negotiation.
\end{itemize}

\section{Situation Monitoring and Delay Prediction} 
\label{sec:datastreams}

\subsection{Situation Monitoring}

\paragraph{Local situation monitoring}

The system uses data from the couriers' smartphone sensors (GPS sensor, rotation sensor, and acceleration sensor) to gain insights into the courier's situation.
For each active courier, the courier agent maintains an up-to-date situation vector.
The situation vector $\vec{s}_i = (s^0_i, s^1_i, ..., s^m_i)$ of courier $i$ contains $m$ features that provide a meaningful summary of the courier's sensor data.
The features have to be chosen by the system designer.
Their choice depends on the specific trigger policy and prediction approach being used and the available data.
Helpful features may include:
the courier's average speed of the last 5 minutes,
the maximal speed of the last 5 minutes,
the number of stops (speed=0) of the last 10 minutes,
notable deviations from the average speed,
sudden accelerations or breaks, etc.
In our evaluation, we use primarily features derived from GPS traces, since the most suitable data sets include GPS data but unfortunately lack data from more diverse sensors.

The features of the situation vector are usually aggregations and abstractions from the low-level sensor data, which makes them more meaningful and stable.
Such aggregations and abstractions can easily be constructed with Complex Event Processing (CEP)~\cite{luckham2001}.
CEP is a rule-based approach for analyzing low-level data streams.
CEP consists of an engine that observes the atomic events of the incoming data stream and uses event processing rules to derive higher-level complex events.
The event processing rules are written in a dedicated event processing language that contains language features to aggregate events observed within certain time windows and to extract complicated, time-dependent event patterns.
CEP has been used in agent-based systems before to extract higher-level situational knowledge from low-level data streams~\cite{doetterl2019,ranathunga2013,ranathunga2012,ziafati2013}.

In our architecture, the higher-level events generated by the CEP engine are used to update the features of $\vec{s}$.
Performing situation monitoring locally on the courier's smartphone reduces the computational load that is placed on the LPA.
Instead of sending each individual sensor event, which would produce many messages the LPA would have to process, only $\vec{s}$ is sent to the LPA whenever $\vec{s}$ changes significantly.
The courier agent continuously updates the courier's situation vector $\vec{s}$, whereby only significant changes in $\vec{s}$ are communicated to the LPA.
Changes in $\vec{s}$ of any degree are communicated to the trigger policy.

\paragraph{Global situation monitoring}
The LPA gathers the situation vectors it receives from the courier agents to keep informed about the couriers' positions and to judge their suitability to take over delivery tasks.

\subsection{Trigger Policy}
Whenever $\vec{s}$ is updated in the courier agent, the trigger policy decides whether to initiate a task transfer.
We apply the following simple policy.
We use the features of $\vec{s}_i$ to estimate a success probability $\sigma_{i,p}$ that courier $i$ can terminate the delivery task $p$ on time.
Whenever $\sigma_{i,p}$ falls below a certain threshold value, the trigger policy initiates a transfer attempt.

\subsection{Delivery Delay Prediction}
\label{sec:delay-prediction}

To estimate $\sigma_{i,p}$, a Hoeffding Tree (HT)~\cite{domingos2000} is used.
HTs are decision trees that can handle the challenging characteristics of data streams.

Conventional decision trees are batch learners that are trained on the whole training data set to learn connections between features and labels.
As data streams are potentially infinite, batch learning becomes infeasible.
Instead, an incremental learner is required that can incorporate new training examples as they arrive over time.

HTs can operate on infinite data streams as they do not store any training examples; in each of their nodes they only maintain a small set of sufficient statistics over the data to decide when to split the node.

\paragraph{Feature vector}

For both learning and prediction, we construct a feature vector $\vec{x}_{i,p}$, which contains for courier $i$ and task $p$ the following features:
\begin{itemize}
\item The remaining distance to the parcel destination
\item The remaining time to the delivery deadline
\item The courier's average speed of the last 5~minutes
\item The courier's maximal speed of the last 5~minutes
\end{itemize}
As the current location of $i$ is known, the distance that $i$ has to travel to deliver $p$ can be computed as $D(l_i, l_p) + D(l_p, d_p)$ where $D(\cdot)$ is the distance function and $l_i$, $l_p$, and $d_p$ are the geographic positions defined in Section~\ref{sec:problem}.
The average and maximal speed are directly taken from $\vec{s}_i$.
If further information is contained in $\vec{s}_i$ that would improve predictions, it can be included into $\vec{x}_{i,p}$.
In our evaluation, we use the mentioned features as these can be derived from the data that is available to us.
The prediction we want to obtain is a probability for each of the two labels $\{ \text{delay}, \text{no-delay} \}$.

\paragraph{Prediction}
To predict whether courier $i$ can deliver $p$ without delay, $\vec{x}_{i,p}$ is passed to the HT.
The HT uses the attribute values of $\vec{x}_{i,p}$ to pass it through the tree.
At each node, the node test determines which edge to follow down the tree.
When a leaf node is reached, the information of the leaf node is used to determine the predicted label.
We use Naive Bayes in the leaf nodes as described by Gama et al.~\cite{gama2006} to compute the label probabilities. 
$HT(\vec{x}_{i,p}) = \sigma_{i,p}$ where $\sigma_{i,p}$ is the estimated probability that $i$ can deliver $p$ in time.

\paragraph{Learning}
To train the HT, we monitor the parcel deliveries.
While a courier $i$ is performing delivery task $p$, in fixed intervals we construct their feature vector $\vec{x}_{i,p}$.
Let $X_{i,p}$ denote the set of these feature vectors.
As soon as the outcome of $p$ is known (delay or no delay), we can use this outcome as the label to train the HT.
Therefore we select one feature vector randomly from $X_{i,p}$ and use it to train the HT.
We use a single sample from $X_{i,p}$ to reduce the impact of training samples gathered from deliveries with long durations.

\paragraph{Evaluation}
In Section~\ref{sec:delay-prediction-eval}, we perform an experimental evaluation that shows the predictive performance of the HT.

\section{Transfer Negotiation}
\label{sec:agreements}

In this section, we discuss three issues regarding the transfer negotiation:
How does the LPA compute transfer suggestions?
Which negotiation protocol do the agents use to find an agreement which transfers to perform?
And under which conditions do two couriers accept the LPA's transfer suggestions?

\subsection{Computing Transfer Suggestions}
When the LPA receives a transfer request for delivery task $p$, it uses its knowledge about the active couriers' situations to select a suitable substitute.
The LPA's objective is to maximize the probability that $p$ arrives on time.
For each courier $i$ that is currently near $p$, the LPA determines $\sigma_{i,p}$.
In line with its objective, the LPA holds preferences over the couriers.
For two couriers $i, j$, the LPA prefers $i$ over $j$ if $\sigma_{i,p} > \sigma_{j,p}$; we say $j$ is dominated by $i$.
The LPA ranks the couriers according to its preferences and discards those that are dominated by the current deliverer, to avoid transfers that worsen the current assignment.
Let $\mathcal{R}$ denote that list (ordered set) of couriers who are not dominated by the current deliverer.

\subsection{Negotiation Protocol}

If $\mathcal{R}$ is empty, no viable transfer is possible and a transfer has to be re-attempted at a later point in time.
If $\mathcal{R}$ is not empty, the LPA's preferred choice is the first courier of the list and the LPA attempts to organize a transfer to this courier.
Let $i$ denote the deliverer and $j$ the first courier in $\mathcal{R}$.

If the LPA had dictatorial power, it would transfer $p$ from $i$ to $j$.
Because both $i$ and $j$ are autonomous agents, the LPA holds no such power over them.
A transfer only takes place when both couriers agree to the transfer.

To attempt a transfer, the LPA offers $p$ to $j$ and asks for their bid $b_j$ (with $b_j \geq 0$).
Bid $b_{j}$ is the payment that $j$ is willing to pay to the current deliverer $i$ to obtain $p$.
If $j$ is not interested in $p$, they respond with $b_{j}=0$.

\begin{itemize}
\item
    If $b_{j}=0$, courier $j$ is removed from $\mathcal{R}$ and the procedure is repeated for the new first courier in $\mathcal{R}$.
\item
    If $b_j>0$, courier $j$ is presented (together with their bid $b_{j}$ and temporal distance $\delta_{j}$) to the deliverer $i$.
    If $i$ agrees to the transfer, $p$ is transferred to $j$.
    If $i$ rejects the transfer, $j$ is removed from $\mathcal{R}$ and the search continues with the new first courier in $\mathcal{R}$.
\end{itemize}

If $\mathcal{R}$ is empty, no viable transfer could be found and a transfer has to be re-attempted later.

\subsection{Transfer Acceptance}
\label{sec:transfer-acceptance}

Under which conditions do the deliverer $i$ and candidate $j$ agree to the transfer?

\subsubsection{Candidate decision}
Candidate $j$ accepts $p$ if $\mathit{U}_{j}(j, p) > 0$ according to the behavior model presented in Section~\ref{sec:problem}.
For $j$, accepting the transfer is the same as accepting an un-assigned parcel.
The courier has to travel to pick up the parcel (either from its initial location or from the deliverer's location) and then deliver it to its destination.
As the only difference, $j$ can effectively gain $r_p - b_j$ instead of $r_p$, as from the reward $r_p$ that is paid out by the system the bid $b_{j}$ flows to the deliverer.

\subsubsection{Deliverer decision}
The deliverer's decision depends on two factors: the bid $b_{j}$ they receive from $j$ and the time they have to wait until $j$ comes by to pick up the parcel if a physical handover is necessary.
To assist the deliverer in their decision, the LPA sends the deliverer, besides $b_j$, the temporal distance $\delta_{j}$ between $i$ and $j$.
The temporal distance $\delta_{j}$ is the estimated time that $j$ will need to visit $i$ and is computed by dividing the distance between $i$ and $j$ through $j$'s current speed.

With this information available, the utility that the deliverer $i$ derives from the (re-)assignment of $p$ to $j$ is defined as follows:
$$\mathit{U}_{i}(j, p) = b_{j} - \mathit{K}(\delta_{j})$$
where $\mathit{K}(\delta_{j})$ denotes the costs (``lost time'') that the deliverer has to endure for awaiting the arrival of $j$.
If the deliverer has picked up the parcel, the substitute $j$ will take $\delta_{j}$ minutes to come by and pick up the parcel.
If the deliverer has not picked up the parcel yet, no physical handover has to be organized and $\mathit{K}(\delta_{j}) = 0$.

The deliverer $i$ accepts the transfer of $p$ to $j$ if the utility derived from the transfer is greater than the utility of finishing the delivery themself:
$$ U_{i}(j,p) > U_{i}(i, p)$$

\section{Evaluation}
\label{sec:evaluation}

We evaluate the stream-based delivery delay prediction of Section~\ref{sec:delay-prediction} and the effectiveness of the whole approach.
As far as possible we use real smartphone GPS data from the BiciMAD data set.

\subsection{GPS Data Set}

The BiciMAD data set\footnote{https://opendata.emtmadrid.es/Datos-estaticos/Datos-generales-(1) (Accessed: 2020-01-29)} contains GPS data from the users of the bike sharing system of Madrid.
Users of the system go to a service station, pick up a bike, use it for short-distance travel within the city, and return it at any of the service stations.
The bike sharing system records data each minute of every trip, including geographic position and speed.
In our experiments, we use data from two dates (15 January and 16 January 2019) with roughly $175,000$ GPS events in total.
For privacy, the GPS events do not carry exact timestamps, but the hour of the day and its occurrence time relative to the (obscured) timestamp of the start of the trip.
As for our experiments we need exact timestamps, we distribute all trips that appeared in the same hour randomly across that hour.
We use this data set because it contains real GPS data from a crowd-based system where users appear, move, and disappear within an urban city center, which is a strong similarity to the crowdshipping system that we envision.
For each of the two days, the data set counts about $8,500$ users (about $17,000$ in total).
In our experiments, we assume that the crowdshipping system can attract a similar number of participants as the bike sharing system and that the participants move by bike.

\subsection{Evaluation of Delivery Delay Prediction}
\label{sec:delay-prediction-eval}
In this section, we perform some explorative experiments for the delay prediction based on smartphone sensor data.
Our goal is not to achieve state-of-the-art performance for arrival time prediction based on smartphone sensor data, but rather to illustrate its potential and usefulness for our main goal: the prevention of delivery delays.
\subsubsection{Experimental Setup}
To evaluate the delivery delay prediction with Hoeffding Trees, we use the GPS data set and for each trip inject an artificial random deadline that lies between 1 and 30~minutes after the trip start.
After each GPS event, we try to predict whether the user will finish their trip within this artificial deadline with the approach of Section~\ref{sec:delay-prediction}.
During the learning process, we compute the prequential error \cite{gama2013}.
When a trip is completed we use the situation vectors that were computed during that trip to predict the delivery outcome.
After each prediction, we compare it to the actual outcome to count the true positives, true negatives, false positives, and false negatives.
Then the Hoeffding Tree is trained with the situation vectors and the actual outcome before the next prediction is made.

\subsubsection{Experimental Results}
\begin{figure}
    \centering
    \includegraphics[width=1.0\linewidth]{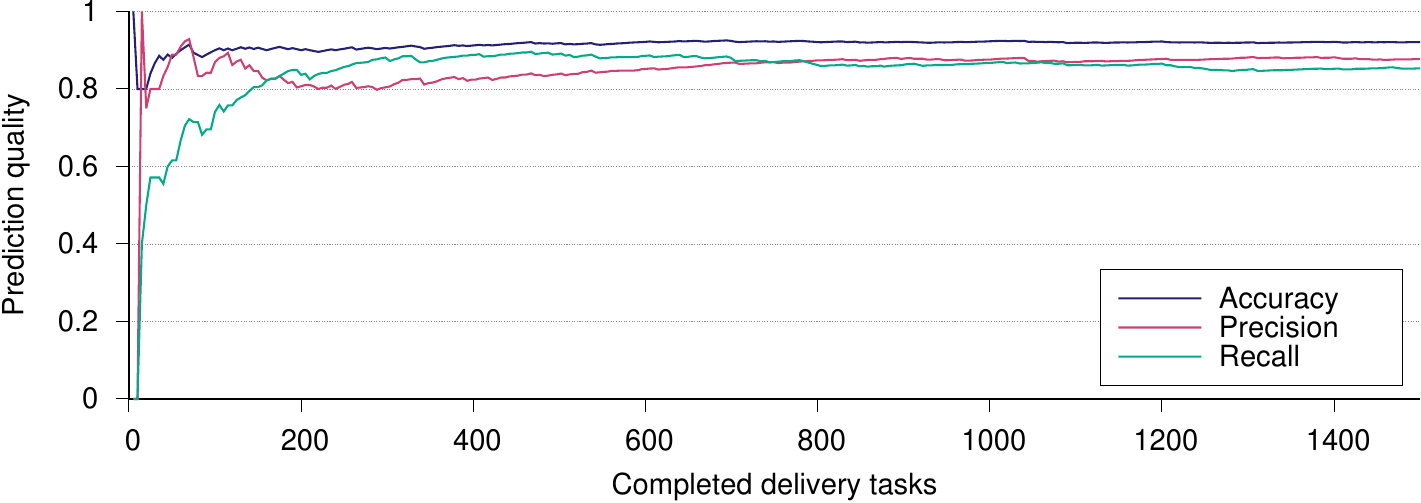}
    \caption{Delay prediction based on smartphone sensor data}
    \label{fig:experiment-vfdt}
\end{figure}
Fig.~\ref{fig:experiment-vfdt} shows the error curves for the three metrics accuracy, precision, and recall.
After few observations, the accuracy reaches about $91\%$.
Over time the precision and recall converge to $88\%$ and $85\%$ respectively.
These values have been achieved by using the default configuration of the scikit-multiflow library \cite{montiel2018}.

\subsection{Evaluation of Delivery Delay Prevention}
\label{sec:evaluation-delay-prevention}
To evaluate the effectiveness of our delay prevention approach, we perform experiments with a crowdshipping simulation.
The experiments are supposed to demonstrate that our approach can effectively prevent delivery delays that would occur without our system.
We evaluate the effectiveness of our approach in different simulation scenarios.

\subsubsection{Crowdshipping Simulation}
We programmed a crowdshipping simulator, that simulates the appearance of delivery tasks in a geographical area.
The simulator also simulates the appearance, movement, and task acceptance behavior of the couriers.
When a new delivery task occurs, it is assigned to the nearest courier who is willing to accept the task.
To simulate the courier appearance and movement before task acceptance, we use the GPS data of the BiciMAD data set.
When a courier accepts a task, they move with constant speed at the direct route to pick up the parcel, then to deliver the parcel, and finally to their original destination as indicated by the data set.
Having accepted a task, incidents can occur that reduce the courier's travel speed to a lower value, which usually provokes a delivery delay.
Incidents can occur with a fixed probability each minute after task acceptance, i.e. the longer a delivery continues the higher becomes the likelihood of an incident.
For each task, the simulator reports the outcome, i.e. whether it was delivered with a delay.

\subsubsection{Experimental Setup}
\begin{table}
    \caption{Simulation setup}
    \centering
    \begin{tabular}{|rc|c|c|}
\hline
        \textbf{Simulation parameter} &  \textbf{Scenario 1} & \textbf{Scenario 2} & \textbf{Scenario 3} \\ \hline
        Delivery area &  \multicolumn{3}{c|}{Center of Madrid, radius 1.5km} \\  
        Parcel reward &  \multicolumn{3}{c|}{EUR 7.0} \\  
        Parcel deadline &  \multicolumn{3}{c|}{30 minutes} \\  
        Penalty for delay &  \multicolumn{3}{c|}{EUR 7.0} \\  
        Travel costs per kilometer & \multicolumn{3}{c|}{EUR 3.0} \\  
        Waiting costs per minute &  \multicolumn{3}{c|}{EUR 0.5} \\  
        Default courier speed &  \multicolumn{3}{c|}{5.0 m/s} \\  
        Courier speed after incident &  \multicolumn{3}{c|}{0.3 m/s} \\  
        Courier prediction error &  \multicolumn{3}{c|}{$[-15, +15]$ minutes} \\
        Trigger threshold &  \multicolumn{3}{c|}{80\%} \\
        Delivery tasks per hour &  50 & 50 & 100 \\  
        Incident probability&   5\% & 10\% & 5\% \\ \hline
\end{tabular}
    \label{tab:simulation}
\end{table}

We use the simulation setup shown in table~\ref{tab:simulation}.
The simulated operating area is a circle with a radius of 1.5km in the center of Madrid.
We use synthetically generated delivery tasks\footnote{The data is available for download:\\http://sw-architecture.inform.hs-hannover.de/files/ECAI2020.zip}, which appear on both simulated days within 8:00 and 20:00 o'clock and have random origins and destinations within the operating area.
Whenever a new task occurs, we assign it to the nearest courier who accepts it.
The couriers accept tasks according to the behavior model of Section~\ref{sec:problem}.
For all tasks, we set a fixed reward of $7$~euro, a deadline of $30$~minutes and a delay penalty of $7$~euro.
The reward is intentionally set high to make sure that all tasks get assigned to a courier.
Each courier wants to receive at least $3$~euro for each kilometer of detour.

During delivery, when they are presented with the option to transfer the task to another courier, they behave according to the transfer acceptance model of Section~\ref{sec:transfer-acceptance}.
We assume that the candidates bid their true valuations.
The deliverers want to get compensated with at least $50$~cents per minute of waiting for the parcel handover.

When couriers have a task assigned, they travel with $5$~m/s as long as no incident occurs.
In case of an incident, the courier's speed drops down to $0.3$~m/s.
If the courier has to estimate the parcel utility, they predict their arrival time with a random error between $-15$ and $+15$ minutes.

For the trigger policy, we use a threshold of $80\%$: if a deliverer's probability of terminating a task on time drops below $80\%$, a transfer is attempted.
We test our approach for different delivery demand ($50$ tasks per hour/$100$ tasks per hour) and for different incident probabilities (5\%/10\%).

We compare three different strategies.
\begin{itemize}
    \item \texttt{NOT} ("No transfer"):
        No transfer takes place.
        This strategy is equal to a crowdshipping system without our transfer approach.
    \item \texttt{S-BEST} ("Suggest best"):
        This strategy uses the approach of Section~\ref{sec:agreements}:
        Couriers are ranked and queried whether they want to take over the task; willing couriers are then presented to the deliverer until the deliverer accepts one (or the transfer attempt fails).
    \item \texttt{F-BEST} ("Force best"):
        The best substitute is determined and forced to take over the delivery task.
        This strategy is similar to \texttt{S-BEST}, but without courier autonomy.
        Instead of seeking agreements between couriers, the couriers are instructed to perform the transfer and they abide by these instructions.
        This strategy should be rather effective but is not viable in practice as it conflicts with our assumption of courier autonomy.
\end{itemize}

We simulate parcel deliveries in three different scenarios.
After each completed delivery, we compute the percentage of deliveries that were completed with a delay.
We visualize this percentage value on a timeline to see how this value evolves over time.
Our goal is to show that approach \texttt{S-BEST} can effectively prevent delays (i.e. outcompetes \texttt{NOT}).
Furthermore, we want to quantify the ``price'' of autonomy: how much worse does \texttt{S-BEST} perform compared to \texttt{F-BEST}?

\subsubsection{Experimental Results}

\paragraph{Scenario 1}

In the first scenario, we test our approach with $50$ parcels per hour and an incident probability of $5\%$.
\begin{figure}
    \centering
    \includegraphics[width=1.0\linewidth]{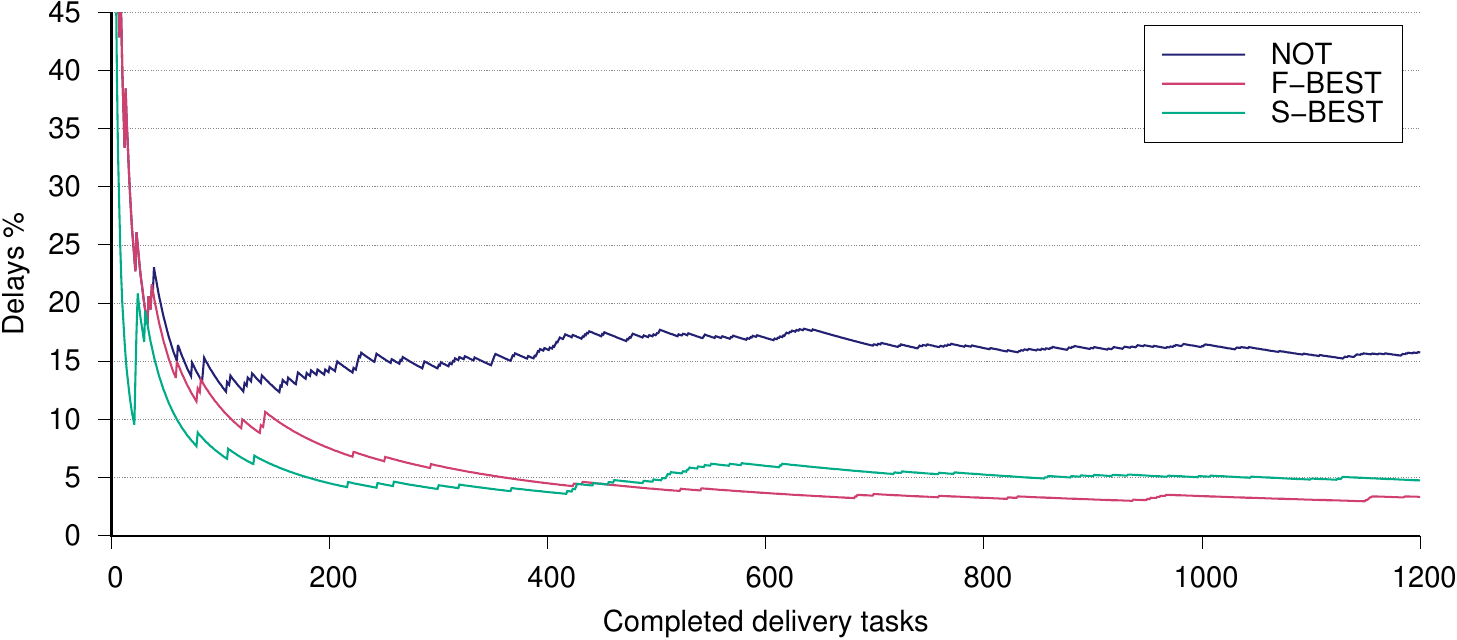}
    \caption{Scenario 1: Delivery delays under different transfer strategies with incident probability 5\% and 50 delivery tasks per hour}
    \label{fig:experiment1}
\end{figure}
Fig.~\ref{fig:experiment1} shows the delivery delays under the different strategies.
Under the \texttt{NOT} strategy, over time the delivery delays converge to around $16\%$.
The \texttt{F-BEST} strategy starts with many delays, as in the beginning the delay predictions are not yet accurate and the strategy forces transfers that are not helpful or even counterproductive.
Over time, when the delay prediction becomes more accurate, the delays converge towards $3\%$.
The \texttt{S-BEST} strategy, somewhat surprisingly, outcompetes \texttt{F-BEST} in the beginning phase (until about 420 parcels), before it converges towards $4.75\%$.
This effect can be explained with the courier autonomy:
in the beginning, the \texttt{F-BEST} strategy enforces counterproductive transfers that result in delays.
At the same time, under \texttt{S-BEST} the couriers reject obviously ill-advised transfer suggestions, which prevents some of the delays.
Over time, \texttt{F-BEST} can reliably identify the most promising transfers and can enforce them, which results in very few delays.
\texttt{S-BEST} also can identify the most promising transfers but has no means to enforce them.
While \texttt{F-BEST} enforces $562$ transfers, \texttt{S-BEST} can only encourage $227$ transfers, which explains why \texttt{S-BEST} is slightly less effective.

\paragraph{Scenario 2}

\begin{figure}
    \centering
    \includegraphics[width=1.0\linewidth]{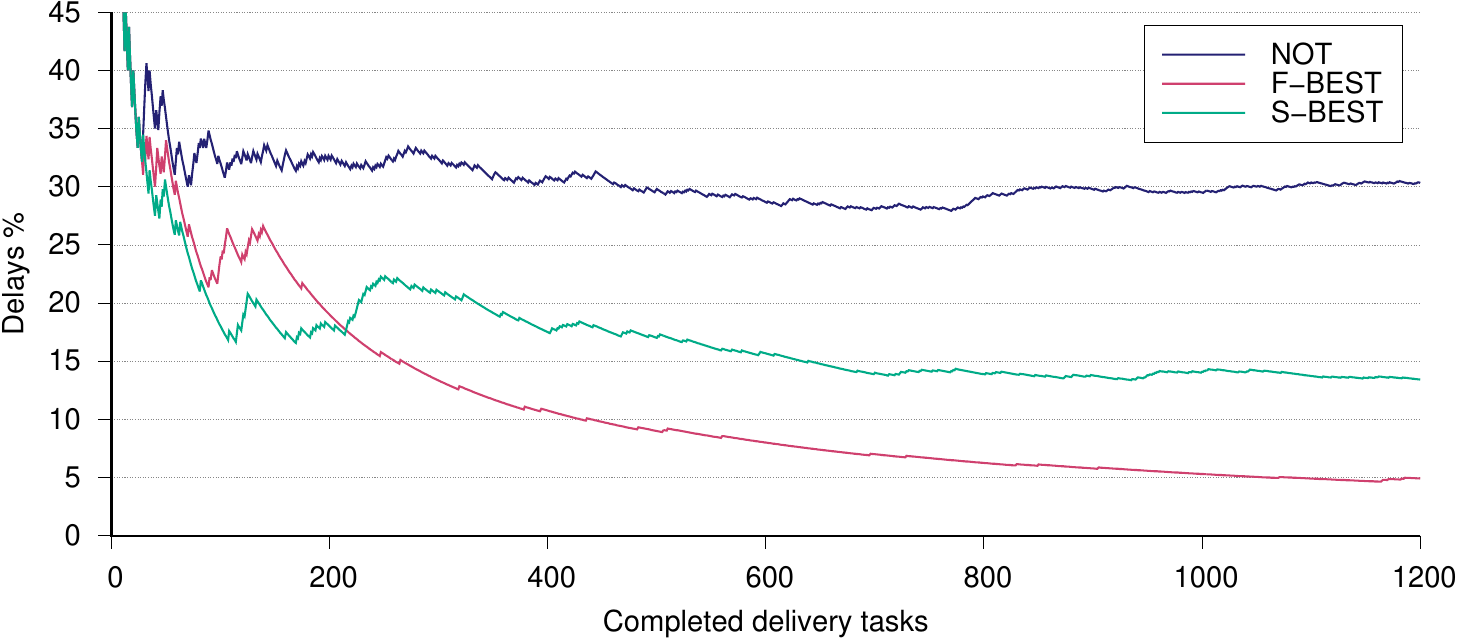}
    \caption{Scenario 2: Delivery delays under different transfer strategies with incident probability 10\% and 50 delivery tasks per hour}
    \label{fig:experiment2}
\end{figure}

As a second experiment, we test how the different strategies perform if we double the incident probability.
Fig.~\ref{fig:experiment2} shows the delivery delays for an incident probability of 10\% and $50$ parcels per hour.
As expected, all three strategies perform worse.
Under the \texttt{NOT} strategy, the delays converge to around $30\%$.
The \texttt{F-BEST} strategy, after the initial phase required for learning, converges towards $5\%$.
The \texttt{S-BEST} strategy converges towards $13\%$.
(Again, \texttt{S-BEST} shortly dominates \texttt{F-BEST} in the initial phase.)
While \texttt{S-BEST} does not reach the low values of \texttt{S-BEST}, it still performs clearly better than \texttt{NOT}.
With such a high incident probability, transfers become less effective as the substitutes who shall take over a task will also be affected by incidents on their way to pick up the parcel from the deliverer.

\paragraph{Scenario 3}
Finally, we perform a third experiment for which we double the number of delivery tasks from $50$ to $100$ and reduce the incident probability back to $5\%$.
\begin{figure}
    \centering
    \includegraphics[width=1.0\linewidth]{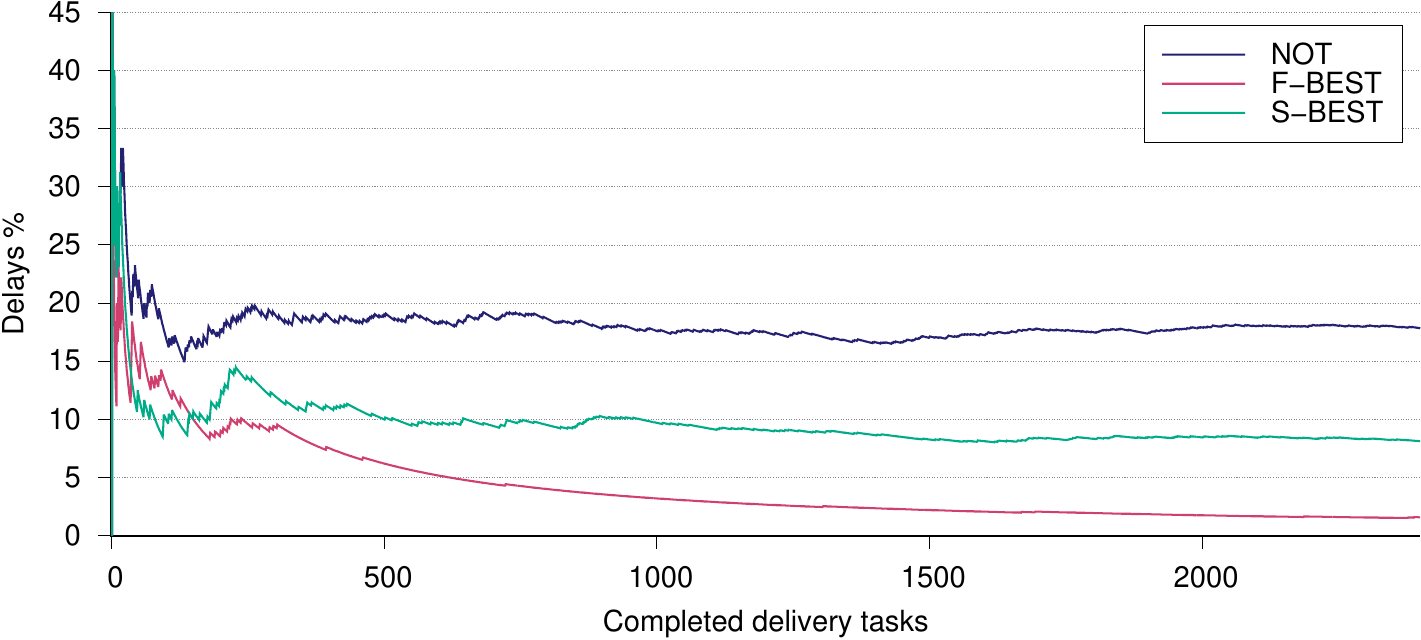}
    \caption{Scenario 3: Delivery delays under different transfer strategies with incident probability 5\% and 100 delivery tasks per hour}
    \label{fig:experiment3}
\end{figure}
As can be seen in Fig.~\ref{fig:experiment3}, \texttt{NOT} converges towards $18\%$, \texttt{F-BEST} converges towards $1.6\%$, and \texttt{S-BEST} converges to around $8\%$.
We compare the outcomes with the outcomes of scenario 1.
By doubling the delivery tasks, only a small increase in delivery delays (from $16\%$ to $18\%$) is observed for the \texttt{NOT} strategy.
For all tasks, a willing deliverer can be found in our crowd.
The \texttt{F-BEST} strategy uses the increased number of delivery tasks to improve its delay prediction and performs even slightly better.
At all times, there is a sufficient number of couriers available to which a transfer can be enforced.
Therefore, \texttt{F-BEST} does not struggle under the increased number of delivery tasks.
\texttt{S-BEST} performs a little worse under the increased number of delivery tasks.
The more delivery incidents occur at the same time, the harder it gets for the approach to find a willing courier who has not already accepted to help with another task.

We conclude from the results of the three scenarios that through purposeful task transfers based on accurate delay predictions many delays can be prevented (\texttt{F-BEST} and \texttt{S-BEST} prevent many delays that occur under the \texttt{NOT} strategy).
The crowd autonomy constrains the space of possible transfers.
Therefore, it is not surprising that our agreement approach (\texttt{S-BEST}) performs slightly worse than enforced transfers (\texttt{F-BEST}).
However, in practice forced transfers are not an option as we hold no dictatorial power over the volunteer couriers of the crowd.

\section{Conclusion}
\label{sec:conclusion}

Traditional parcel delivery methods struggle with the ``last mile'':
The delivery of parcels from hubs to the final recipient is costly and challenging for time-sensitive deliveries.
Crowdshipping is an alternative delivery method that leverages the crowd.
However, it is hard to achieve on-time delivery when deliveries are delegated to the crowd because it is dynamic and consists of autonomous individuals.

In this paper, we have made several contributions towards on-time parcel delivery with crowds.
We have presented an agent-based crowdshipping architecture consisting of courier agents and a logistics provider agent.
Courier autonomy is one of the key differences to fleets of professional drivers who are employees who can be managed by the logistics provider.
Therefore it is crucial to model and treat the couriers as autonomous agents that have the final say over their personal decisions.

To be able to prevent delivery delays proactively, before they occur, we have incorporated prediction capabilities into the logistics provider agent.
In our agent-based system, the agents perform data stream processing on smartphone sensor data to maintain situational knowledge about the active couriers and deliverers.
The logistics provider uses data stream mining to derive delivery delay predictions.
The use of smartphone sensor data for delay prediction distinguishes our approach from other agent-based crowdshipping proposals that have been made in the literature.

Our experiments have shown that through accurate delay predictions and informed task transfers many delays can be prevented.
We have observed that the autonomy of the crowd makes it difficult to prevent some delays even if they can be predicted accurately.
Sometimes, no delay-preventing transfer is feasible if the available couriers that would be capable to complete the task in time are not willing to.
For future work, our task transfer approach should be coupled with an advanced payment scheme that allows the logistics provider agent to convince more couriers to follow its transfer suggestions.
(Such a payment scheme has recently been proposed to incentivize autonomous taxi drivers to agree to taxi reassignments~\cite{billhardt2019}.)
If the logistics provider agent had advanced budgeting capabilities, it could potentially forge transfer agreements that can not be achieved with the simple transfer payments that we use in this paper.

Task transfers can require parcels to be handed over from deliverers to substitutes.
The handover can be organized by having the deliverer wait for the substitute to come by to receive the parcel, which is the approach we have used in this paper.
This method might be inconvenient for the deliverer, especially if in a hurry.
As alternatives, the deliverer and the substitute could agree on a meeting point, or the parcel could be temporarily stored in nearby packing station until its delivery is continued by the substitute.

Currently, our approach is limited to one delivery task per courier.
If a courier takes over more than one task, it is harder to predict for each task whether it will be completed in time without asking the courier in which order they intend to perform the deliveries.
Future work should allow multiple tasks per courier as this way potentially redundant trips could be avoided, which would not only help to reduce costs but would also be more sustainable.

Lastly, our transfer approach works well in small spatial operating areas where the crowd is dense.
When the operating area gets larger, the distances between the potential couriers get larger which makes it harder to find capable and willing transfer candidates.
Future work should investigate how task transfers can be scaled to sparse crowds.

\bibliographystyle{ecai}
\bibliography{bibliography}

\end{document}